\documentclass{article}
\usepackage{spconf,amsmath,graphicx}
\usepackage{hyperref}
\usepackage{commath}
\usepackage{amsfonts}
\usepackage{xcolor}
\usepackage{algorithm}
\usepackage{algpseudocode}
\usepackage[mathscr]{euscript}
\usepackage{enumitem}
\usepackage{booktabs}
\usepackage{multirow}
\usepackage{graphicx}
\usepackage{array}
\usepackage{setspace}

\title{Learned Image Compression with Text Quality Enhancement}
\twoauthors
 {Chih-Yu Lai\sthanks{Work done during internship at Microsoft.}}
	{Massachusetts Institute of Technology\\
	Electrical Engineering and Computer Science\\
	Cambridge, MA, USA\\
        \href{mailto:chihyul@mit.edu}{chihyul@mit.edu}}
 {Dung Tran and Kazuhito Koishida}
	{Microsoft Corporation\\
	Applied Sciences Group\\
	Redmond, WA, USA\\
        \{\href{mailto:dung.tran@microsoft.com}{dung.tran},\href{mailto:kazukoi@microsoft.com}{kazukoi}\}@microsoft.com}

\begin{document}
%
\maketitle

\begin{abstract}
Learned image compression has gained widespread popularity for their efficiency in achieving ultra-low bit-rates. Yet, images containing substantial textual content, particularly screen-content images (SCI), often suffers from text distortion at such compressed levels. To address this, we propose to minimize a novel text logit loss designed to quantify the disparity in text between the original and reconstructed images, thereby improving the perceptual quality of the reconstructed text. Through rigorous experimentation across diverse datasets and employing state-of-the-art algorithms, our findings reveal significant enhancements in the quality of reconstructed text upon integration of the proposed loss function with appropriate weighting. Notably, we achieve a Bjontegaard delta (BD) rate of -32.64\% for Character Error Rate (CER) and -28.03\% for Word Error Rate (WER) on average by applying the text logit loss for two screenshot datasets. Additionally, we present quantitative metrics tailored for evaluating text quality in image compression tasks. Our findings underscore the efficacy  and potential applicability of our proposed text logit loss function across various text-aware image compression contexts.
\end{abstract}

\section{Introduction}
\label{sec:intro}
Learned image compression, primary focusing on lossy compression, has gained significant traction in deep learning \cite{liu2023learned,hu2021learning,cheng2020learned,codevilla2021learned,PPR:PPR644584, jamil2022learningdriven,mentzer2020learning}. Recent advances, leveraging entropy models \cite{ballé2017endtoend,ballé2018variational,cheng2020learned,liu2023learned,Duan_2023_WACV}, have even surpassed the established Versatile Video Coding (VVC) standard \cite{xie2021enhanced,he2022elic,Duan_2023_WACV}. However, while excelling with natural images, these methods often archive suboptimal reconstructions for screen content images (SCIs) due to susceptibility to coding artifacts and distinctive dissimilarities in luma histograms between SCIs and natural images\cite{9897719,heris2023multitask}.

To enhance compression performance for SCIs, several efforts have emerged in the recent literature. Guo et al. introduced a causal contextual prediction model, outperforming VVC at low bitrates but exhibiting limitations at higher bitrates \cite{Guo_2022}. Wang et al. proposed an end-to-end compression scheme that incorporates the transform skip concept tailoring to screen content characteristics, yielding superior results compared to existing non-SCI-specific compression models \cite{9897719}. Heris et al. contributed a deep learning-based codec designed specifically for SCIs, demonstrating efficacy in input reconstruction and synthetic/natural region segmentation tasks \cite{heris2023multitask}. Other related tasks such as super-resolution focused on SCIs have also been addressed to tackle real-word challenges \cite{10.1145/3589963}. Nonetheless, these methods primarily focus on the visual aspects, overlooking textual content within images. Therefore, there remains a critical need for methods explicitly accounting for reconstructed text quality. 

\begin{figure*}
    \centering
    \includegraphics[width=\textwidth]{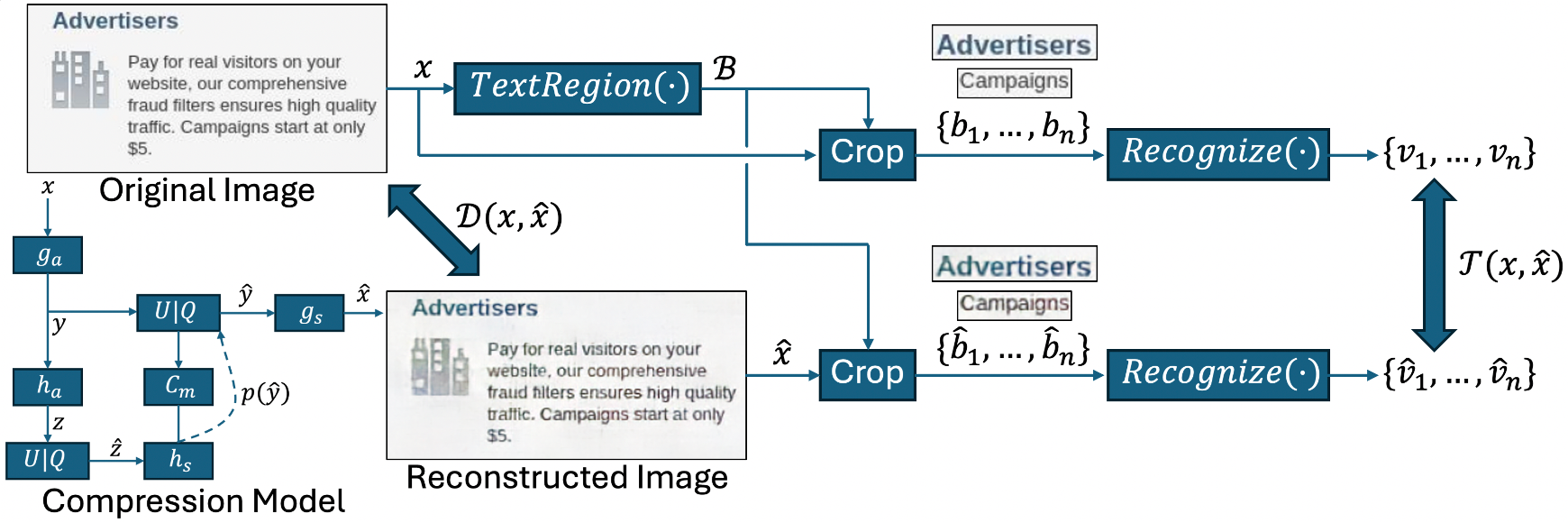}
    \caption{High-level training scheme for text-aware image compression: The coordinates for all text regions inside the original image are extracted as $\mathscr{B}$ and used to crop the original ($x$) and reconstructed ($\hat{x}$) images into two lists of cropped text regions, $\{b_1,...,b_n\}$ and $\{\hat{b}_1,...,\hat{b}_n\}$. The text is recognized to obtain lists of logits $\{v_1,...,v_n\}$ and $\{\hat{v}_1,...,\hat{v}_n\}$, and then compared to calculate the text logit loss ($\mathscr{T}(x,\hat{x})$). During backpropagation, the loss gradient is reflected in the weights of the compression model.}
    \label{fig:scheme}
\end{figure*}

Addressing the challenge of text quality in learned image compression, we introduce a novel text logit loss designed to measure the disparity in textual fidelity between the original and reconstructed images. Our method leverages Optical Character Recognition (OCR) and scene text recognition models to detect and recognize text within images, upon which the proposed loss function is constructed. By minimizing this loss, we aim to enhance the perceptual similarity of text between the two images. Fig. \ref{fig:scheme} illustrates the primary scheme for learned image compression with text quality enhancement. Our contributions are summarized as follows:
\begin{enumerate}[leftmargin=*]
    \item We propose a method for text-aware learned image compression. To our knowledge, this is the first work that explicitly focuses on enhancing text quality within learned image compression.
    \item Furthermore, we introduce the plug-and-play text logit loss $\mathscr{T}(x,\hat{x})$ capable of operating with various state-of-the-art models. By adjusting weight values, our approach allows for control over the quality of reconstructed text.
    \item Finally, we present Bjontegaard delta for Character Error Rate (BD-CER) and Bjontegaard delta for Word Error Rate (BD-WER) metrics for quantitatively assess the fidelity of reconstructed text across diverse image compression methods, offering a comprehensive evaluation framework for benchmarking the performance of compression techniques. 
\end{enumerate}

\section{Methods}
\label{sec:methods}

Consider a total loss function $\mathscr{L}$ for learned image compression. This total loss function comprises various components, including those measuring dissimilarities between the original and reconstructed images (e.g. distortion loss, $\mathscr{D}$), as well as those capturing the compressed image size (e.g. rate loss, $\mathscr{R}$). In this context, we introduce the text logit loss $\mathscr{T}$ which serves as an additional component incorporated into the total loss function to characterize disparities in textual fidelity.

\subsection{Text Logit Loss}
\label{ssec:loss_func}

The Text Logit Loss, $\mathscr{T}(x, \hat{x})$, is calculated from the original image, $x$, and the reconstructed image, $\hat{x}$. $\mathscr{T}(x, \hat{x})$ quantifies the dissimilarity between $x$ and $\hat{x}$ in terms of character-based logits, where a list of logits is extracted from an image by identifying, cropping, and recognizing its text regions.

Formally, let's assume we have a function, $TextRegion(\cdot)$, that returns a list of bounding boxes, ${B_1, B_2, ..., B_n}$, for all individual text regions inside the ground truth image, $x$,
\begin{gather}
    \mathscr{B} = \{ B_1, B_2, ..., B_n \} = TextRegion(x)
\end{gather}

\noindent where each $B_i$ can be defined by the coordinates of its top-left and bottom-right corners. We utilize $B_i$ as a function to crop any input image based on these coordinates,
\begin{gather}
    b_i = B_i(x), \; \hat{b}_i = B_i(\hat{x})
\end{gather}

\noindent where $b_i, \hat{b}_i \in \mathbb{R}^{h_i \times w_i \times 3}$ represent the $i$-th cropped text region of $x$ and $\hat{x}$, respectively, with height $h_i$ and width $w_i$. This approach ensures that the same regions between $x$ and $\hat{x}$ are compared. Additionally, it addresses the challenge of extracting meaningful text regions from the output of an untrained compression model.

Assume we have another function, $Recognize(\cdot)$, that recognizes and classifies each individual character inside a given text region,
\begin{gather}
    v_i = Recognize(b_i), \; \hat{v}_i = Recognize(\hat{b}_i)
\end{gather}


\noindent where $v_i, \hat{v}_i \in \mathbb{R}^{T \times S}$ represent the text logit outputs, with $T$ being the maximum character length and $S$ being the size of the possible character set \cite{bautista2022scene}. In general, $v_i$ and $\hat{v}_i$ are associated with the unnormalized prediction values for each individual character in $b_i$ and $\hat{b}_i$, where a higher value corresponds to a higher probability of being that specific character. We then obtain the text logit loss by calculating the mean squared value of the text logit difference between $x$ and $\hat{x}$.
\begin{gather}
    \mathscr{T}(x, \hat{x}) = \frac{1}{n} \sum^{n}_{i=1}{\norm{v_i - \hat{v}_i}^2}
\end{gather}

\subsection{End-to-end Training Flow}
\label{ssec:train_flow}

The text logit loss can be incorporated into any image compression model without modifying any model-specific architecture. This can be observed from the fact that $\mathscr{T}(x, \hat{x})$ only requires the original and reconstructed images as its inputs. For entropy-based compression models, we add the weighted value of $\mathscr{T}(x, \hat{x})$ to the total loss function,
\begin{gather}
\label{eq:total_loss}
    \mathscr{L} = \mathscr{R}(\hat{y}) + \mathscr{R}(\hat{z}) + \lambda \cdot \mathscr{D}(x, \hat{x}) + \kappa \cdot \mathscr{T}(x, \hat{x})
\end{gather}

\noindent where $\mathscr{R}(\hat{y})$ and $\mathscr{R}(\hat{z})$ are the rate losses for the compressed codes $\hat{y}$ and $\hat{z}$, $\mathscr{D}(x,\hat{x})$ is the distortion loss between $x$ and $\hat{x}$, $\lambda$ is the weight parameter for the distortion loss, and $\kappa$ is the weight parameter that controls the importance of the text logit loss. Since $\mathscr{T}(x, \hat{x})$ is calculated solely from the cropped text regions, its direct impact is confined to the quality of text reconstruction in those specific regions. If $TextRegion(\cdot)$ and $Recognize(\cdot)$ are fixed, $\mathscr{T}(x, \hat{x})$ serves as a parameter-free loss function. This implies that our method does not introduce any additional degree of freedom but rather guides the training process toward text-aware compression.

Note that $\mathscr{B}$, $b_i$, and $v_i$ ($i \in \{1, 2, ..., n\}$) can be pre-calculated for each ground truth image before training. In such cases, $TextRegion(\cdot)$ doesn't need to be applied during training since $\mathscr{B}$ can be directly used for cropping $\hat{x}$, and $Recognize(\cdot)$ only needs to be applied to $\hat{x}$ during forward propagation. This approach significantly reduces training time at the cost of using extra memory to save $\mathscr{B}$, $b_i$, and $v_i$. In practice, $\mathscr{L}$ is averaged over a batch of images $\mathbf{x} = \{x_1, x_2, ..., x_m\}$ to balance between accuracy and speed.

We provide the algorithm for calculating (\ref{eq:total_loss}) over $m$ images in Algorithm \ref{alg:train}, given $TextRegion(\cdot)$, $Recognize(\cdot)$, and some entropy compression model $Model(\cdot).$ Note that our method can be applied not only to entropy models but also to more general compression models, as long as backpropagation can be applied to the reconstructed output.

\begin{algorithm}
\caption{Total Loss Function}\label{alg:train}
\begin{algorithmic}
\Require $\mathbf{x} = \{x_1, x_2, ..., x_m\}, \lambda, \kappa$
\State $\mathscr{L} \gets 0$
    \For{$k \gets 1$ to $m$}
        \State $\hat{x}_k, \hat{y}_k, \hat{z}_k \gets Model(x_k)$
        \State $\mathscr{B}_k = \{B_{k1},B_{k2},...,B_{kn}\} \gets TextRegion(x_k)$
        \State $\mathscr{T}_k \gets 0$
        \For{$i \gets 1$ to $n$}
            \State $b_{ki},\hat{b}_{ki} \gets B_{ki}(x_k),B_{ki}(\hat{x}_k)$
            \State $v_{ki},\hat{v}_{ki} \gets Recognize(b_{ki}),Recognize(\hat{b}_{ki})$
            \State $\mathscr{T}_k \gets \mathscr{T}_k + \norm{v_{ki}-\hat{v}_{ki}}^2$
        \EndFor
        \State $\mathscr{T}_k \gets \mathscr{T}_k / n$
        \State $\mathscr{L} \gets \mathscr{L} + \mathscr{R}(\hat{y}_k) + \mathscr{R}(\hat{z}_k) + \lambda\cdot \mathscr{D}(x_k, \hat{x}_k) + \kappa\cdot \mathscr{T}_k$
    \EndFor
    \State $\mathscr{L} \gets \mathscr{L} / m$
\State Return $\mathscr{L}$
\end{algorithmic}
\end{algorithm}

\begin{figure*}[ht]
    \centering
    \includegraphics[width=\textwidth]{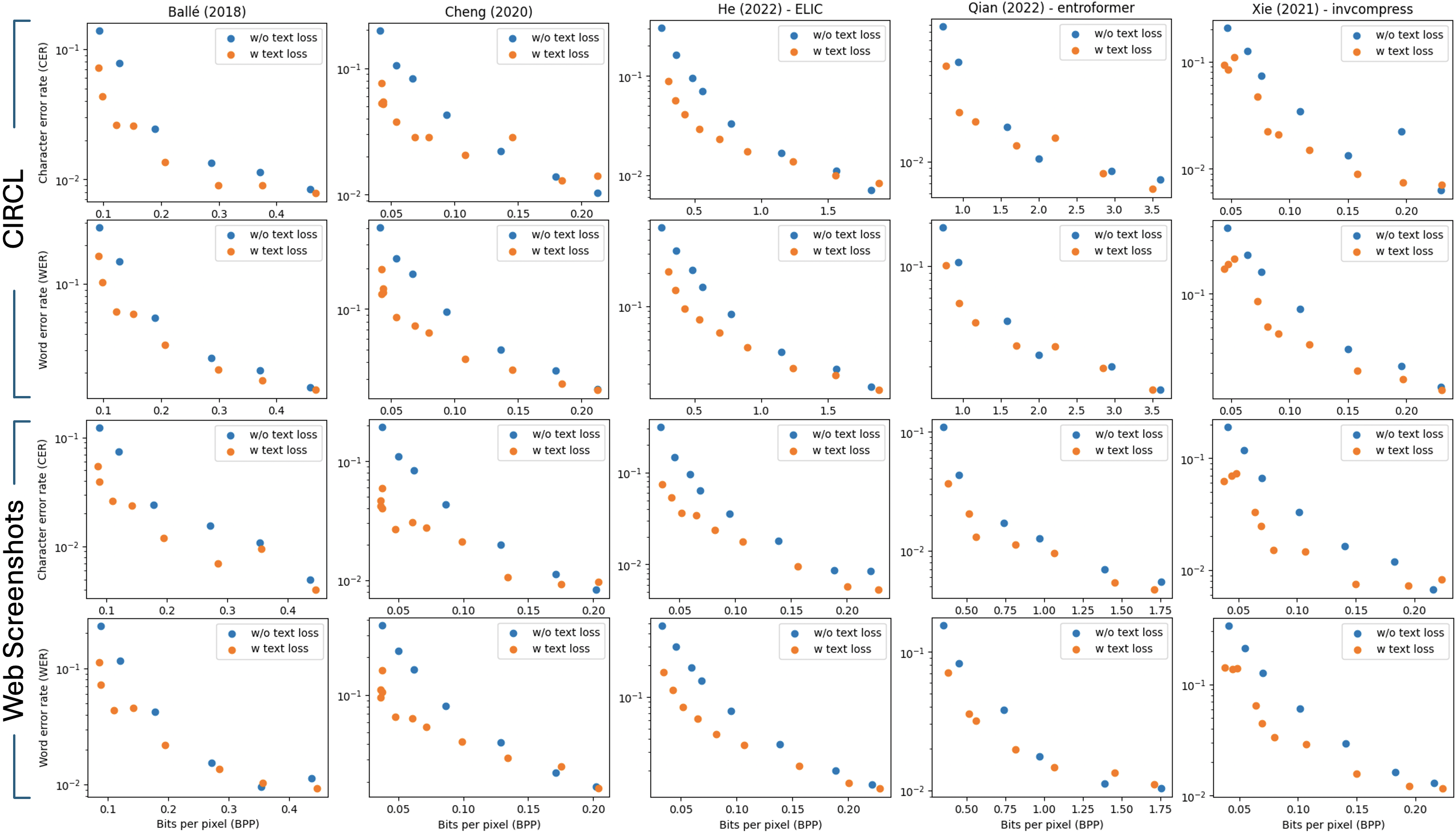}
    \vspace{-15pt}
    \caption{Character error rate (CER) vs Bits per pixel (BPP) and Word error rate (WER) vs BPP for the CIRCL and Website Screenshot datasets. Results with and without using the text logit loss are shown. $\kappa = 0.1$ for all experiments. Generally, the CER/WER when using the text logit loss are lower at the same BPP compared to not using the text logit loss. This is more significant when the BPP is lower.}
    \vspace{-5pt}
    \label{fig:cerwer}
\end{figure*}

\vspace{-10pt}
\section{Experiments}
\label{sec:experiments}

We evaluate our approach by comparing the outcomes with and without using the text logit loss during training. We utilize multiple datasets and state-of-the-art entropy-based learned compression algorithms for this evaluation.

\subsection{Datasets}
\label{ssec:datasets}

Since our proposed method focuses on enhancing text quality for image compression tasks, evaluating our method on natural image datasets is not suitable. Therefore, our primary focus lies on Screen Content Images (SCIs). In particular, we evaluate our method using two screenshot datasets, where one has undergone lossless compression (PNG-based), and the other has undergone lossy compression (JPEG-based).

\begin{itemize}[leftmargin=*]
    \item \textbf{CIRCL Images AIL}: The open-source online screenshot dataset consists of images scraped from onion domain websites by AIL (Analysis Information Leak framework) \cite{CIRCL-AILDS2019}. Images in this dataset are in PNG format (losslessly compressed). All models are trained using a specific subset of the dataset - \verb|folder_0.tar.gz|, even when the models are tested using other datasets. The remaining subset - \verb|folder_1.tar.gz| - is reserved for testing. To adhere to the rule defined in (\ref{eq:eng_retain}) for retaining bounding boxes, we exclude images containing fewer than 5 English text regions during training.
    
    \item \textbf{Roboflow Website Screenshots}: A synthetically generated dataset comprises screenshots from over 1000 of the world's top websites \cite{WebsiteScreenshotsDataset-2024-01-23}. The images in this dataset are in JPG format (lossy compressed). We utilize the test split portion (121 images) for testing. Given that lossy compressed images are commonly encountered, evaluating the generalization ability of using the text logit loss for image compression on this dataset becomes significant.
\end{itemize}

\subsection{Implementation Details}
\label{ssec:implement}

We employ the CRAFT text detector to implement the $TextRegion(\cdot)$ function \cite{baek2019character}. In particular, the pretrained detector is sourced from the open-source EasyOCR library \cite{EasyOCR-2024-01-23}. For the $Recognize(\cdot)$ function, we utilize the PARSeq Scene Text Recognition (STR) model \cite{bautista2022scene} that is pre-trained for recognizing English text. The settings for text region detection and recognition remain unchanged compared to the official version. By applying the same method for calculating the text logit loss, using any recognition model that is pre-trained on other languages will also work.

Since the raw output of CRAFT doesn't differentiate between English and other languages \cite{baek2019character}, we filter the output of PARSeq to determine whether to keep or discard individual text regions, thereby retaining the English text regions. Specifically, if a bounding box $b$ of some corresponding text logit $v$ needs to be retained for training, it should satisfy the inequalities
\begin{gather}
\label{eq:eng_retain}
    \mathrm{median}(v) \geq m_{min}, \;\; \mathrm{stdev}(v) \leq \sigma_{max}
\end{gather}

In practice, we set $m_{min} = 14.2$ and $\sigma_{max} = 2$. For each setting, we train our model for an initial 20 epochs. If the results do not converge, an additional 20 epochs of training are performed. The training batch size is set to 8, and the learning rate is set to $10^{-4}$. When the text logit loss is added to the total loss, the resulting BPP (bits per pixel) will vary for the same model. To ensure a fair comparison, we use different ranges of $\lambda$ with and without the text logit loss, aiming for approximately the same resulting range of BPPs. For each experiment, $\kappa$ is set to 0.1, and $\lambda$ is set to a value between 0.0001 and 0.01. For Ballé (2018) \cite{ballé2018variational} and Cheng (2020) \cite{cheng2020learned}, we implement them using the CompressAI PyTorch library \cite{begaint2020compressai} (specifically, \verb|bmshj2018_hyperprior| and \verb|cheng2020_attn|). For all other algorithms, we use the implementation code from online github repositories.

\begin{figure*}[ht]
    \centering
    \includegraphics[width=\textwidth]{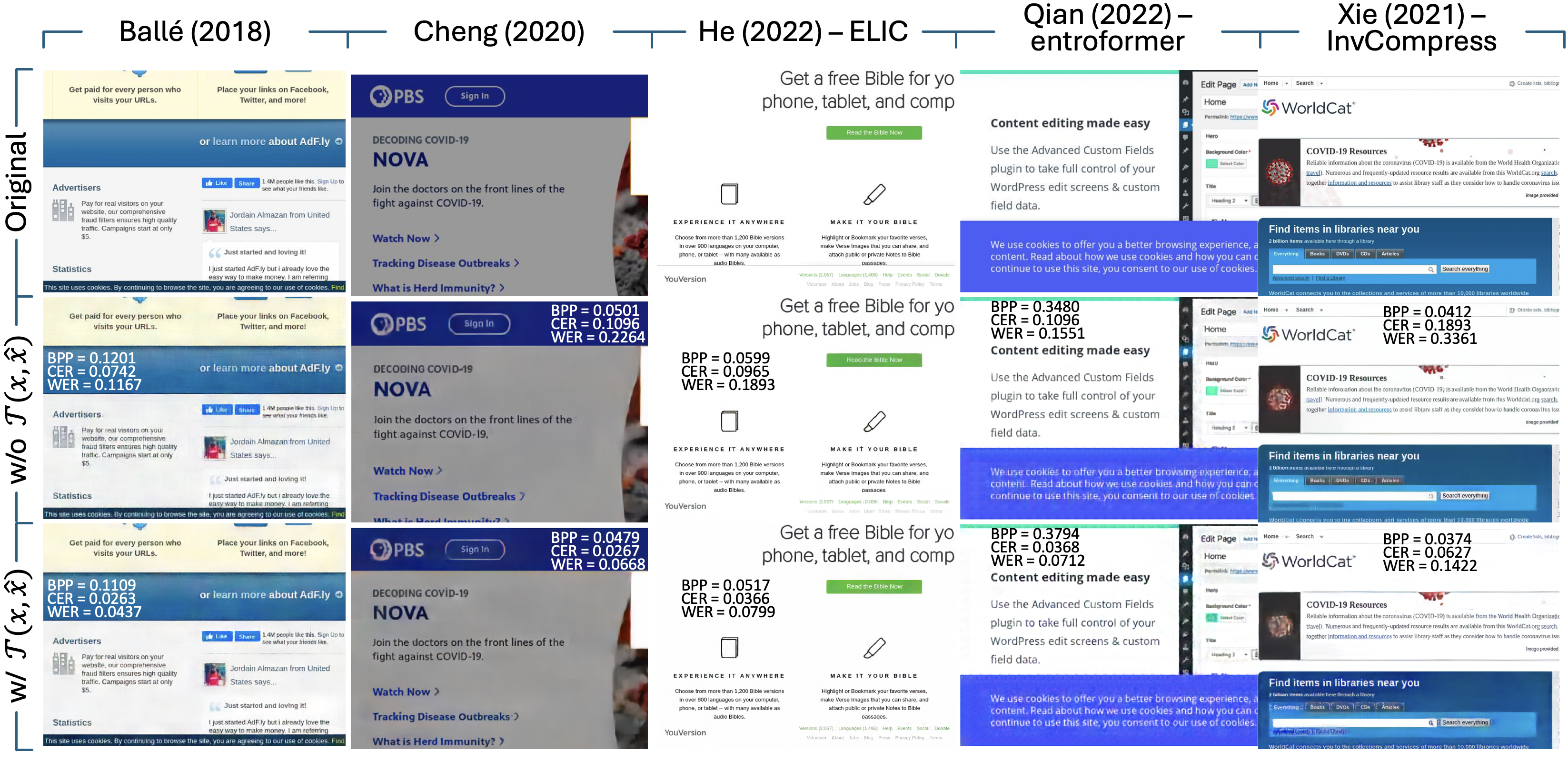}
    \vspace{-15pt}
    \caption{Original and reconstructed images from five screenshots in the WebScreenshots dataset. 'w/o $\mathscr{T}(x,\hat{x})$' denotes training without using the text logit loss, while 'w/ $\mathscr{T}(x,\hat{x})$' indicates training with the text logit loss. Perceptually, the reconstructed text without the text logit loss appears blurrier and more distorted. For example, in the bottom-left figure (Ballé 2018 with text logit loss), the text at the bottom line is easily readable, whereas in the center-left figure (Ballé 2018 without text logit loss), it is harder to read.}
    \label{fig:visualization}
    \vspace{-5pt}
\end{figure*}

\subsection{Quantitative Results}
\label{ssec:quantresult}

\noindent\textbf{Metrics.} We use the Character Error Rate (CER) and Word Error Rate (WER) to characterize the performance of our experiments \cite{10.1145/3476887.3476888,rajesh2023hwrcnet,8270041}.
\begin{gather}
    \mathrm{CER} = \frac{S_c+D_c+I_c}{S_c+D_c+C_c}, \;\; \mathrm{WER} = \frac{S_w+D_w+I_w}{S_w+D_w+C_w}
\end{gather}

Here, $S_c$ and $S_w$ represent the number of substitutions, $D_c$ and $D_w$ denote the number of deletions, $I_c$ and $I_w$ stand for the number of insertions, and $C_c$ and $C_w$ indicate the number of correct items. These values are calculated either for individual characters (subscript $c$) or words (subscript $w$). Since each valid text region corresponds to one word, and we always perform text recognition on the corresponding regions in the reconstructed image, $D_w = 0$ and $I_w = 0$ in our experiments. This implies that WER $\leq 1$ for every image. However, such a restriction does not apply to CER. The CERs and WERs are first calculated for each individual image and then averaged over all images.

To measure the average BPP reduction at the same CER/WER, we use the Bjontegaard delta (BD) rate for quantification. The BD Rate measures the BPP reduction offered by an algorithm feature while maintaining the same quality, as assessed by objective metrics \cite{barman2024bjontegaard}. Similar to the calculation of BD-PSNR \cite{Bjntegaard2001CalculationOA}, in order to measure the average CER/WER reduction at the same BPP, we define the BD-CER and BD-WER as the percentage changes in CER and WER while maintaining the same BPP when comparing a target method to a reference. \\

\noindent\textbf{Results.}
We compare the results of five entropy model-based algorithms \cite{ballé2018variational,cheng2020learned,he2022elic,qian2022entroformer,xie2021enhanced}. Fig. \ref{fig:cerwer} plots the evaluation results across all models and screenshot datasets for learned image compression. The comparison includes results when using the text logit loss (\ref{eq:total_loss}) and when not using the text logit loss. CER and WER exhibit a high correlation throughout all experiments. Across all algorithms and datasets, the use of the text logit loss generally leads to a decrease in CER and WER at the same BPP. In other words, the curve is shifted downwards when the text logit loss is additionally applied. For the same algorithm, this difference is more significant in terms of ratio at a lower BPP. Indeed, setting a higher $\lambda$ provides the models with greater memory capacity to compress the image, allowing it more flexibility to balance between saving text-related information and other details.

Furthermore, we note that additionally applying the text logit loss generally results in a higher BPP and lower CER/WER for a fixed $\lambda$. This trend is discussed in section \ref{ssec:changekappa}. Even when trained on the CIRCL training dataset, which is lossless, the algorithms can still perform well on the Web Screenshots dataset with lossy compression images, regardless of the presence of the text loss. This highlights the generality of the entropy models and the text loss.

Table \ref{tab:bd_rate} presents the Bjontegaard delta rate (BD Rate) for CER and WER in each experiment, with the reference method being the original algorithm without applying the text logit loss. Table \ref{tab:bd_cer_wer_psnr} presents the BD-CER, BD-WER, and BD-PSNR for the same set of experiments. Although, there is a slight drop in BD-PSNR after adding the text logit loss, we can clearly observe a significant performance improvement in terms of CER and WER at the same BPP. This demonstrates the effectiveness of the text logit loss when applied to screenshot images.

\begin{table}[ht]
\renewcommand{\arraystretch}{1.3}
\huge
\centering
\caption{BD Rate (\%) comparing the CER and WER between using ($\kappa=0.1$) or not using (reference) the text logit loss. The BPP decreases at the same CER/WER after adding the text logit loss.}
\label{tab:bd_rate}
\resizebox{\columnwidth}{!}{%
\begin{tabular}{@{}cccccccc@{}}
\toprule[1.5pt]
\textbf{Dataset} & \textbf{Metric} & \textbf{Ballé (2018)} & \textbf{Cheng (2020)} & \textbf{He (2022)} & \textbf{Qian (2022)} & \textbf{Xie (2021)} & \textbf{Mean} \\ \midrule
\multirow{2}{*}{CIRCL}           & CER & -16.58 & -36.96 & -36.73 & -24.93 & -35.25 & -30.09 \\
                                 & WER & -20.03 & -36.41 & -34.33 & -17.02 & -33.48 & -28.25 \\
\midrule
\multirow{2}{*}{\shortstack[c]{Website\\Screenshots}} & CER & -30.04 & -46.40 & -40.68 & -17.62 & -41.16 & -35.18 \\
                                 & WER & -24.17 & -40.90 & -39.41 & -12.40 & -22.19 & -27.81 \\ \bottomrule[1.5pt]
\end{tabular}%
}
\end{table}

\begin{table}[ht]
\renewcommand{\arraystretch}{1.3}
\huge
\centering
\caption{BD-CER (\%), BD-WER (\%), and BD-PSNR (\%) between using ($\kappa=0.1$) or not using (reference) the text logit loss. The CER/WER improves and PSNR deteriorates at the same BPP after adding the text logit loss.}
\label{tab:bd_cer_wer_psnr}
\resizebox{\columnwidth}{!}{%
\begin{tabular}{@{}cccccccc@{}}
\toprule[1.5pt]
\textbf{Dataset} & \textbf{Metric} & \textbf{Ballé (2018)} & \textbf{Cheng (2020)} & \textbf{He (2022)} & \textbf{Qian (2022)} & \textbf{Xie (2021)} & \textbf{Mean} \\ \midrule
\multirow{3}{*}{CIRCL} & BD-CER & -34.08 & -22.06 & -30.85 & -7.06 & -49.17 & -28.64 \\
 & BD-WER & -25.70 & -35.52 & -32.16 & -12.18 & -38.76 & -28.86 \\
 & BD-PSNR & -4.07 & -0.92 & -0.77 & -0.06 & -0.99 & -1.36 \\
\midrule
\multirow{3}{*}{\shortstack[c]{Website\\Screenshots}} & BD-CER & -41.33 & -42.05 & -43.16 & -26.52 & -50.33 & -40.68 \\
 & BD-WER & -22.95 & -30.49 & -39.07 & -13.93 & -43.90 & -30.07 \\
 & BD-PSNR & -3.50 & -0.65 & -0.55 & -0.27 & -0.86 & -1.17 \\ \bottomrule[1.5pt]
\end{tabular}%
}
\end{table}

\begin{figure*}[ht]
    \centering
    \includegraphics[width=\textwidth]{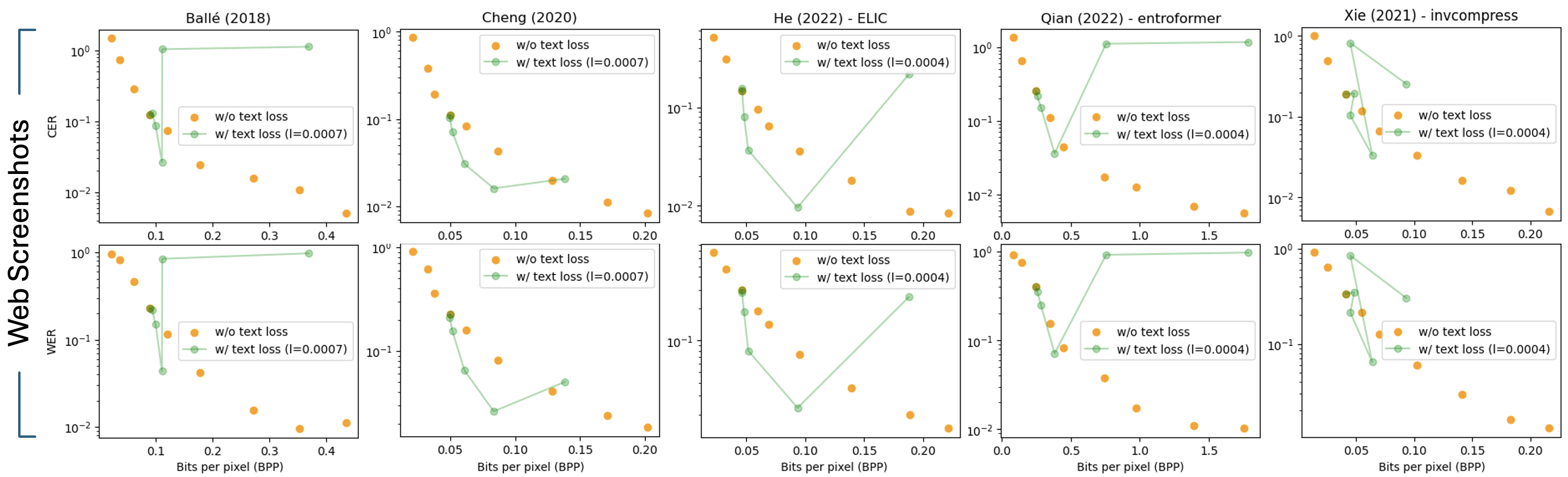}
    \vspace{-15pt}
    \caption{CER and WER vs BPP for the Website Screenshot dataset. The results are presented using the text loss for one specific $\lambda$ with $\kappa \in {0.001, 0.01, 0.1, 1, 10}$, and without using the text loss for $\lambda \in {0.0001, 0.0002, 0.0004, 0.0007, 0.001, 0.002, 0.004, 0.007, 0.01}$. Starting from $\kappa = 0.001$, the CER and WER show a more pronounced drop compared to keeping $\kappa=0$ and increasing $\lambda$, indicating that a small $\kappa$ is useful for relatively increasing text quality. When $\kappa$ is too large, CER and WER become excessively high, which is undesirable.}
    \label{fig:kappas}
    \vspace{-5pt}
\end{figure*}

\subsection{Qualitative Results}
\label{ssec:visualization}

Fig. \ref{fig:visualization} presents visualizations of the original and reconstructed images across different algorithms. Regarding the reconstructed images with or without using text loss, we choose images with the closest BPPs for comparison. Qualitatively, we observe that at the same BPP, the reconstructed images using text loss for training exhibit better text quality. For regions that contain natural images, adding text loss results in similar qualities. This also holds true for text regions that are not in English (data not shown) since the compression model loses focus on non-English text during training. Moreover, since the text loss primarily focuses on the recognized text logits, the colors in each pixel slightly differ more from the text in the original image compared to the reconstructed image without using text loss. Overall, the results show that by using the text logit loss, we can relatively increase the text reconstruction quality while maintaining the same BPP during compression.

\subsection[Effect of kappa on Text Quality]{Effect of $\kappa$ on Text Quality}
\label{ssec:changekappa}

The weight parameter $\kappa$ is crucial to the model during training. In Fig. \ref{fig:kappas}, we plot the training results without using the text logit loss and using the text logit loss for a specific $\lambda$ at different $\kappa$. Note that setting $\kappa=0$ corresponds to not using the text logit loss. We observe that when starting with a small $\kappa$ and gradually increasing it, the CER/WER generally decreases, and the BPP generally increases. Moreover, for a smaller $\kappa$, the decrease in CER/WER is generally larger compared to not using the text logit loss when the increase in BPP are the same.

We can tune $\kappa$ to find a balanced point that achieves a relatively lower CER/WER without an excessive increase in BPP. However, as $\kappa$ becomes larger, there is a risk that the algorithm might not be able to maintain stable training, resulting in an exploding CER/WER. For instance, in the case of ELIC \cite{he2022elic}, if $\kappa=10$, both the CER/WER and BPP will be excessively high. This indicates that we cannot freely increase $\kappa$ to achieve better text reconstruction quality. We need to strike a balance between distortion, rate, and text loss to achieve adequate model performance, with the balancing weight parameters being algorithm-specific to some extent.

\section{Conclusion}
\label{sec:conclusion}

In conclusion, we have proposed a novel text logit loss designed to enhance the quality of reconstructed text in learned image compression. By quantifying the disparity in text between original and reconstructed images, we aimed to minimize this loss, thereby improving the perceptual quality of the reconstructed text. Our experiments across two screenshot datasets and five state-of-the-art compression algorithms demonstrated the effectiveness of our approach. We observed significant improvements in both CER and WER when employing the text logit loss. Additionally, we introduced the BD-CER and BD-WER metrics, highlighting the relative improvement in text quality at the same bitrate. Qualitatively, visualizations of reconstructed images showed clearer and less distorted text when trained with the text logit loss. The investigation on the weight parameter $\kappa$ shows that a small positive $\kappa$ value effectively increases text quality without excessively compromising overall compression performance. Our findings underscore the potential of incorporating text-aware techniques in learned image compression systems, paving the way for future advancements in this field.

\clearpage



\begingroup
\setstretch{0.95}
\bibliographystyle{IEEEbib}
\bibliography{refs}
\endgroup

\end{document}